\documentclass[10pt,twocolumn]{article} 
\usepackage{style}
\usepackage{times}
\usepackage{graphicx}
\usepackage{amssymb}
\usepackage{url,hyperref}
\usepackage{mathtools}
\usepackage[labelfont=bf]{caption}

\begin{document}

\title{Recurrent Control Nets as Central Pattern Generators \\
for Deep Reinforcement Learning}

\author{\textbf{Vincent Liu$^\dagger$ \quad Ademi Adeniji$^{\dagger,*}$ \quad Nathaniel Lee$^{\dagger,*}$ \quad Jason Zhao$^{\dagger,*}$} \\
\texttt{\{vliu15, ademi, natelee, jzhao23\}@stanford.edu} \\ \\
\textbf{Mario Srouji$^\dagger$} \\
\texttt{msrouji@stanford.edu} \\ \\
$^\dagger$Department of Computer Science, Stanford University \\
$^*$Equal Contribution \\ \\
}

\maketitle
\thispagestyle{empty}

\begin{abstract}
Central Pattern Generators (CPGs) are biological neural circuits capable of producing coordinated rhythmic outputs in the absence of rhythmic input. As a result, they are responsible for most rhythmic motion in living organisms. This rhythmic control is broadly applicable to fields such as locomotive robotics and medical devices. In this paper, we explore the possibility of creating a self-sustaining CPG network for reinforcement learning that learns rhythmic motion more efficiently and across more general environments than the current multilayer perceptron (MLP) baseline models. Recent work introduces the Structured Control Net (SCN), which maintains linear and nonlinear modules for local and global control, respectively \cite{scn}. Here, we show that time-sequence architectures such as Recurrent Neural Networks (RNNs) model CPGs effectively. Combining previous work with RNNs and SCNs, we introduce the Recurrent Control Net (RCN), which adds a linear component to the, RCNs match and exceed the performance of baseline MLPs and SCNs across all environment tasks. Our findings confirm existing intuitions for RNNs on reinforcement learning tasks, and demonstrate promise of SCN-like structures in reinforcement learning.
\end{abstract}

\section{Introduction}
The ability to model rhythmic outputs given arrhythmic inputs has significant implications in neuroscience, robotics, and medicine. CPG-modeling networks are a vastly important yet little understood region of reinforcement learning. Existing baselines for locomotive tasks fail to provide rhythmic intuitions to reinforcement learning agents, although its incorporation demonstrate training and reward improvements over previous baselines \cite{scn}.

In reinforcement learning, the agent makes decisions based off a policy, which is a mapping from each state to an action. The model, during training, learns the mapping that will maximize reward. The standard policy network used when modeling locomotive tasks is the multilayer perceptrons  \cite{mlp2, es1, mlp3}. OpenAI uses an MLP of 2 hidden layers and 64 hidden units as its baseline model (MLP-64). This simple neural network learns to model tasks of moderate complexity, but fail to generate rhythmic output without rhythmic input \cite{scn}.

Instead, recent papers propose architectures which split the policy network into linear and nonlinear components \cite{resnet, scn}. \cite{scn} introduces the Structured Control Net, which performs better than the standard MLP across many environments. The SCN architecture is comprised of a linear module for local control and a nonlinear module for global control (in \cite{scn}, an MLP-16), the outputs of which sum to produce the policy action. We use this baseline (SCN-16) for our benchmarks.

\begin{figure}
\centering
   \includegraphics[width=80mm,scale=0.5]{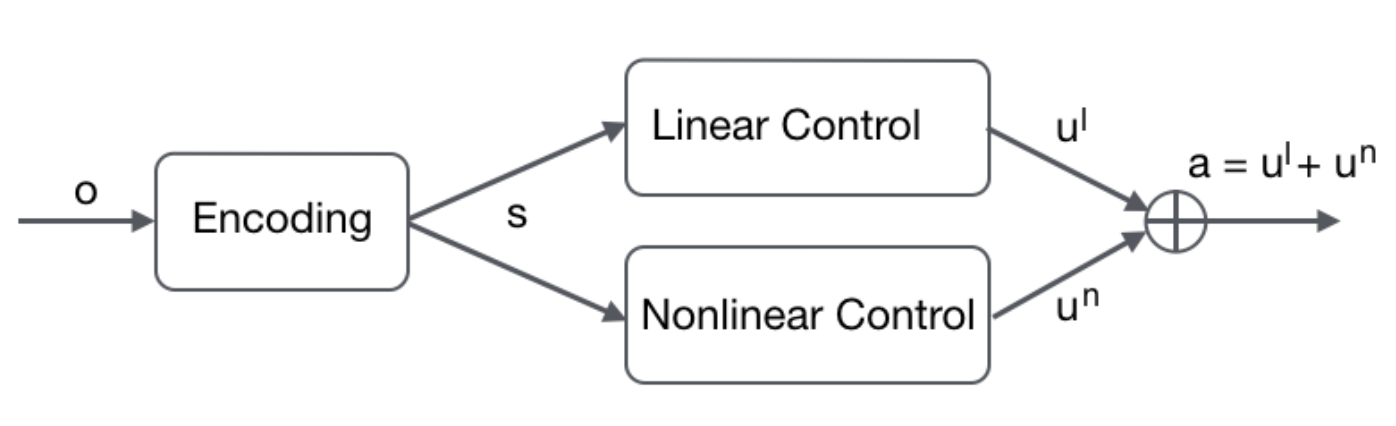}
   \caption{\textmd{Structured Control Net architecture}}
   \label{fig:scn}
\end{figure}

There exists literature that explores the possibility of applying Recurrent Neural Networks (RNNs) to the central pattern generation task \cite{rnn2, rnn1}. RNNs maintain a hidden state that is updated at each timestep, which allows RNNs to produce context-informed predictions based on previous inputs from past sequences \cite{rnn3}. The simplest (vanilla) RNN allows inputs and previous hidden states to flow between time-states freely and can provide more context to the action policy at the current timestep. The RNN itself has many limitations, including loss of information to long time sequences, vanishing and exploding gradients, and complexity of paralleliztion. \cite{rnn3}. We also explore the efficacy of variations to the vanilla RNN intended to mitigate these shortcomings including Long Short-Term Memory RNNs \cite{lstm1, lstm2} and Gated Recurrent Units \cite{gru} and provide results on the detrimental effect of RNN complexity on our RL environments. 

In this paper, we combine the intuition behind SCNs and RNNs. We adopt the separation of linear and nonlinear modules, which has been shown to improve performance by learning local and global interactions with RNNs as our nonlinearity. We experimentally demonstrate that this architecture brings together the benefits of both linear, nonlinear, and recurrent policies by improving training sampling efficiency, final episodic reward, and generalization of learned policy, while learning to generate actions based on prior observations. We further validate our architecture with competitive results on simulations from OpenAI MuJoCo, trained with the Evolutionary Strategies \cite{es1, es2, es3} optimization algorithm.




\section{Related Work}
MLPs have previously been used to attempt RL modeling of rhythmic control tasks. The intuition is that the nonlinear fully-connected architecture is an effective function approximation. Although MLPs are able to generate high episodic rewards on many OpenAI tasks, they converge to locomotive behaviors that are jerky and unintiutive to motion.

\subsection{Structured Control Nets}
\cite{scn} provides evidence that locomotion specific structures than general purpose MLPs can provide motion intuitions to RL environments. Brielfy, the SCN architecture is to separately learn local and global control. These are specific to locomotion tasks: the agent needs to learn global interactions and general patterns of movements specific to a task. \cite{scn} demonstrates improvements only using a MLP nonlinear module. These SCN results follow network intuitions, that since locomotive actions are also heavily dependent on immediate prior actions learning local interactions gives local context necessary to generate subsequent actions. Still, the SCN learns cyclic actions very weakly, as it produces outputs from current observations only. The Recurrent Control Net instead provides deeper context from previous states that better provides context for these locomotive RL policies, allowing the possibility of cyclic policy functions.

\begin{figure}
\centering
   \includegraphics[width=80mm,scale=0.5]{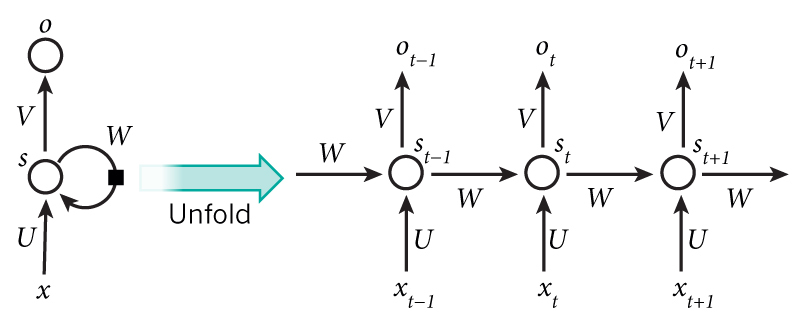}
   \caption{\textmd{Recurrent Neural Network architecture}}
   \label{fig:rnn}
\end{figure}

\subsection{Recurrent Neural Networks}
RNNs model time-sequences well by maintaining a hidden state as a function of priors \cite{rnn3}. Previously, RNNs have been intensively used in natural language processing because sequence prediction is best modeled by including previous context. Therefore, to leverage the added context, RNNs has also been explored loosely in reinforcement learning for quadruped environments. However, RNNs have not been generalized to general locomotive tasks and still remain relatively specific to these quadruped based locomotion tasks \cite{rnn1, rnn2}. Instead of RNNs as policy networks, inclusion of linear-global dynamics by adopting simple RNN as a nonlinearity in SCNs have the potential to exceed existing SCN/MLP Baselines (See Appendix for a comprehensive description of effective RNN architectures).


\begin{figure*}
\centering
\includegraphics[width = 0.23\textwidth]{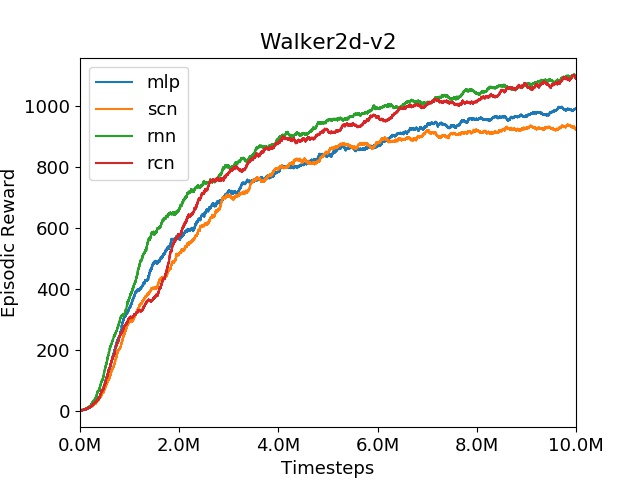} 
\includegraphics[width = 0.23\textwidth]{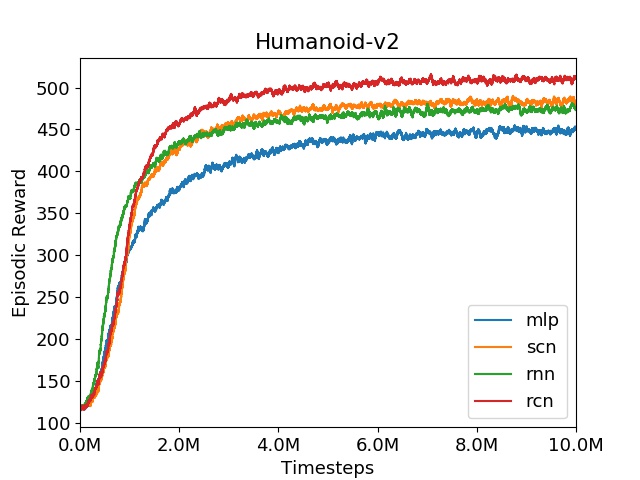}
\includegraphics[width = 0.23\textwidth]{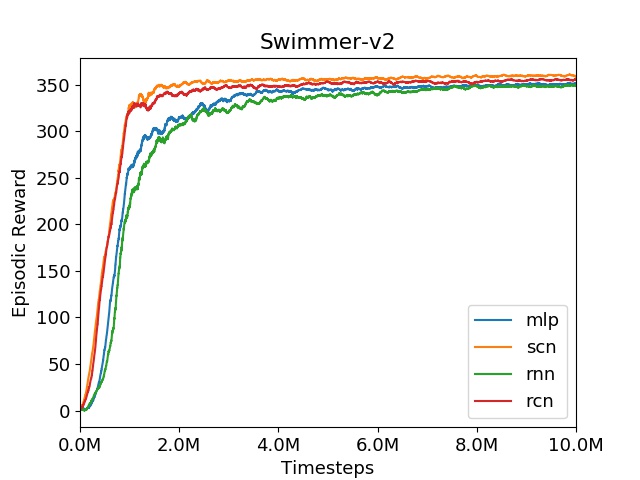}
\includegraphics[width = 0.23\textwidth]{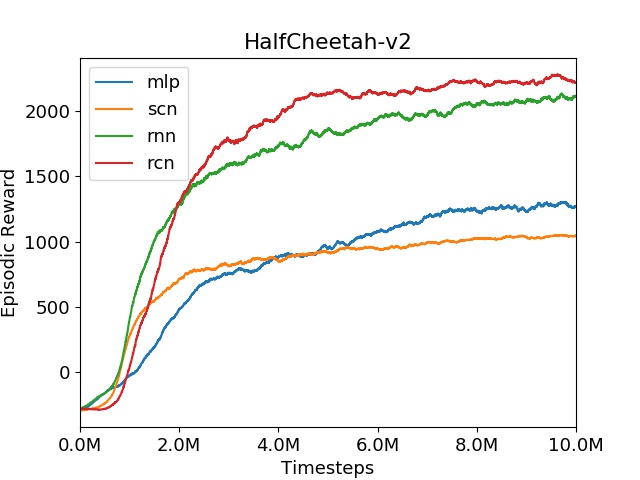}
\caption{Episodic rewards for MuJoCo environments on baselines MLP-64, SCN-16, RNN-32, and RCN-32 using ES optimization. Average of 5 median trials from 10 total trials.}
\label{fig:baselines}
\end{figure*}

\begin{figure}
\centering
\includegraphics[width = 0.23\textwidth]{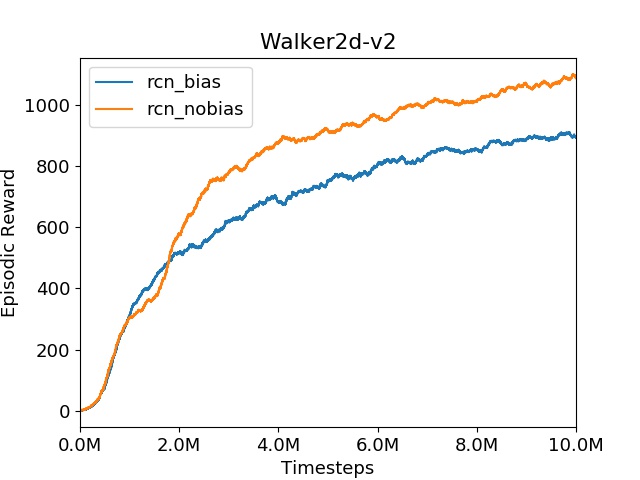} 
\includegraphics[width = 0.23\textwidth]{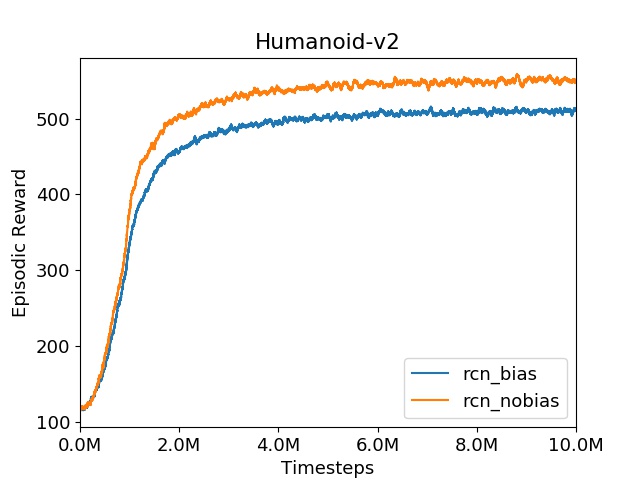}
\includegraphics[width = 0.23\textwidth]{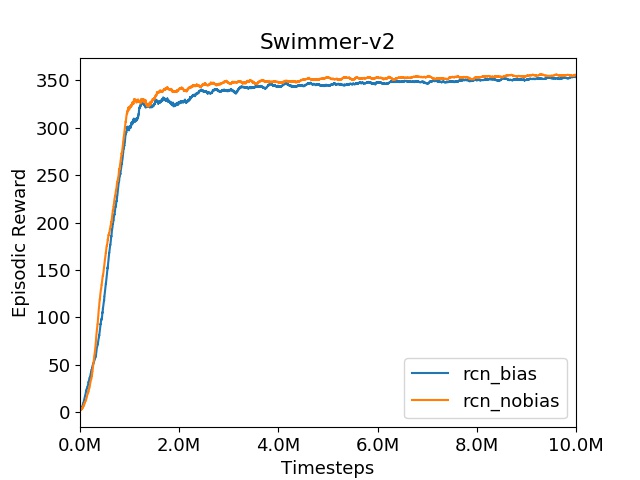}
\includegraphics[width = 0.23\textwidth]{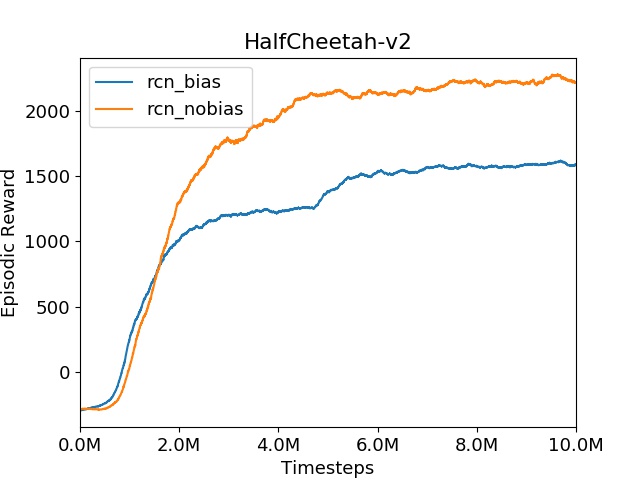}
\caption{Episodic rewards on MuJoCo environments with RCN-32, with and without bias vectors using ES optimization. Average of 5 median trials from 10 total trials.}
\label{fig:rcn-biases}
\end{figure}

\section{Experimental Setup}
To work with locomotive tasks, we use OpenAI Gym \cite{openaigym}, a physics environment for reinforcement learning, and run our models on Multi-Joint dynamics with Contact (MuJoCo) tasks \cite{mujoco}. The Gym environment effectively serves as a wrapper to the MuJoCo tasks.

At each timestep, the Gym environment returns an observation, which encodes the agent's state (i.e. joint angles, joint velocities, environment state). The policy takes this observation as input and effects an action back into the environment. In cyclic fashion, the environment returns to the policy the action reward and subsequent observation. Over many episodes and timesteps, the policy learns how to traverse the environment by maximizing the rewards of its actions.

\section{Recurrent Control Net}
We utilize the findings from \cite{scn} of separate linear and nonlinear modules and design our Recurrent Control Net (RCN) in a similar fashion. The linear module is identical to that of the SCN \cite{scn}, but the nonlinear module is a standard vanilla RNN with hidden size 32. Intuitively, the linear module provides local control while the nonlinear module provides global control. However, unlike the MLP used in SCN-16 \cite{scn}, the RNN learns global control with access to prior information. We use this architecture (RCN-32) as our baseline in experiments.

In our experiments, we compare the RCN-32 to a vanilla RNN of hidden size 32 (RNN-32) to test the efficacy of the extra linear module. To reduce the number of trainable parameters for ES, we remove all bias vectors in all models.

\begin{figure}
\centering
\includegraphics[width = 0.23\textwidth]{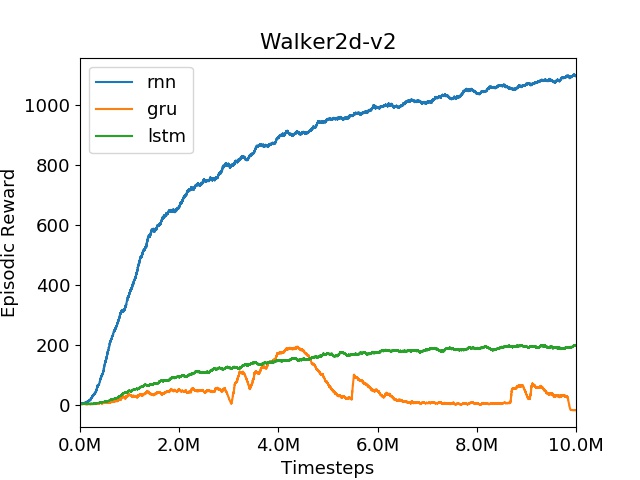} 
\includegraphics[width = 0.23\textwidth]{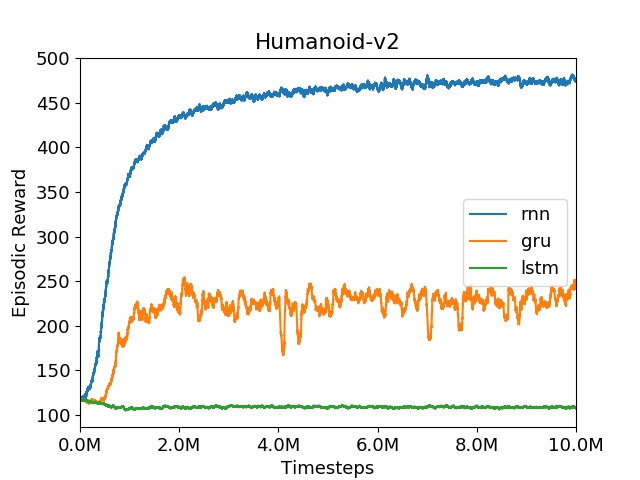}
\includegraphics[width = 0.23\textwidth]{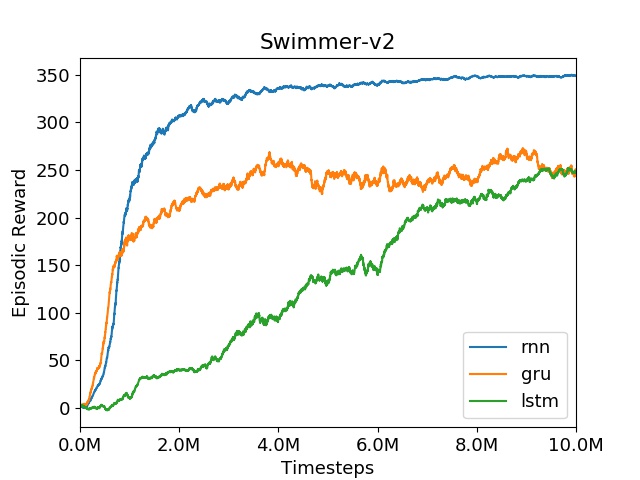}
\includegraphics[width = 0.23\textwidth]{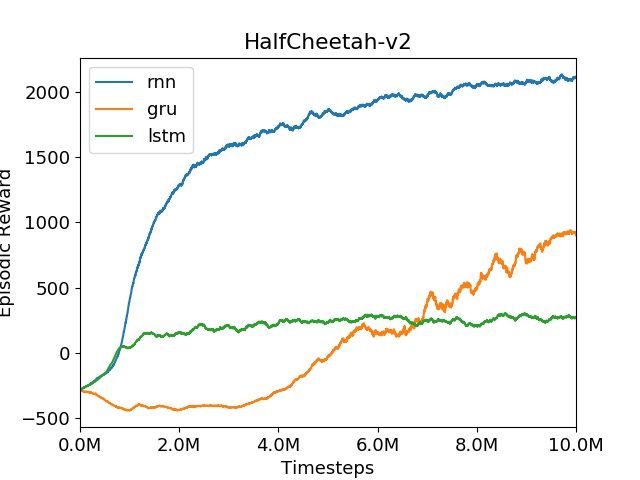}
\caption{Episodic rewards on MuJoCo environments with RNN-32, GRU-32, and LSTM-32 using ES optimization. Average of 5 median trials from 10 total trials.}
\label{fig:rnns}
\end{figure}

\section{Evaluation}
We use OpenAI's MLP-64 model and the SCN-16 outlined in \cite{scn} as the baseline for comparison. We evaluate a model's efficacy by its final reward after 10M timesteps, not by the training curve (rate of convergence). Across all MuJoCo environments, we find that the RNN-32 matches or exceeds both baselines (see Figure \ref{fig:baselines}). We also notice that the RCN-32 consistently improves upon the RNN-32.

From our experimentation with various recurrent structures, we can make several interesting observations. The recurrent structure seems to be inherently conducive to modeling locomotive tasks because its hidden state explicitly encodes past observations, whereas a multilayer perceptron does so implicitly. We also find that the increase in model complexity past a certain threshold is detrimental to ES's randomized training process. Additionally, explicit modeling of linear and global interactions with linear and nonlinear modules consistently improves model performance.

\subsection{Gated Information Flow}
In all our trials with ES optimization, we notice that recurrent architectures with gated information flow (GRUs, LSTMs) struggle in training. We believe that since ES is a random optimizer, it struggles to optimize models with more parameters (see Figure \ref{fig:rnns}). A more complex model introduces more local optima, which may cause ES to converge earlier. Additionally, since MuJoCo tasks are relatively simple and low-dimensional, enhanced memory is unnecessary and learning gates in training only burdens the model.

The ES algorithm is inherently hampered by its gradient-free approach. Because it updates weights with random noise, even slightly more complex models suffer in training. However, with simple architectures, we see early convergence in episodic reward (compared to the same models trained with different algorithms). As is such, we anticipate that GRUs and LSTMs may achieve higher rewards with an optimization algorithm like PPO (Proximal Policy Optimization, a gradient based optimizer that updates parameters through gradient ascent \cite{ppo}).

\subsection{Linear Control}
We notice that the RCN-32 consistently outperforms the RNN-32. This finding is consistent with the proposal in \cite{scn}, which shows the efficacy of introducing a linear component in addition to the nonlinear component (see Figure \ref{fig:baselines}). Intuitively, the linear module learns local interactions while the nonlinear module learns global interactions. As the RNN is a nonlinear network with a layer activation after the hidden state update, the addition of the linear module to the architecture allows it to learn local and global interactions. While the increase in performance is marginal at times, the SCN-16 improves the MLP-64 by similar amounts.

\subsection{Incorporating Biases}
We also experiment with adding biases into the RCN. Doing so immediately decreases performance across all tasks, even dropping performance to below those of the baselines (see Figure \ref{fig:rcn-biases}). We believe that this is because the inclusion of biases burdens the optimizer in training without providing any real value to what the model learns.

\section{Conclusion}
We conclude that RNNs model locomotive tasks well. Furthermore, we also conclude that the separation of linear and nonlinear control modules improves performance. The RCN thus learns local and global control while learning patterns from prior sequences. We also note the detriment of increasing model complexity with information gates, though this is probably due to MuJoCo task simplicity and the ES training algorithm. Because ES updates weights randomly, additional gates create more local optima that ES struggles to overcome.

As we only train with the ES algorithm, it would be interesting to explore the performance of RCNs with an algorithm such as Proximal Policy Gradient. Additionally, recent practices in natural language processing stray away replace recurrent layers and with convolutional layers \cite{ds3}. It would be interesting to see if convolution could replace the RNN module. We hope that this opens up further investigation into the usage of RCNs in these tasks.

\section{Appendix: Recurrent Architectures}
For more context, this section will cover in depth the architectures of Recurrent Neural Networks (RNNs), Gated Recurrent Units (GRUs) and Long Short-Term Memories (LSTMs).

\subsection{Recurrent Neural Network}
The vanilla RNN maintains an internal hidden state to compute future actions, which serves as a memory of past observations. This simple architecture allows all inputs and hidden states to flow freely between timesteps. Standard RNN update equations are below\footnote{$h^{(t)}$, $o^{(t)}$, $x^{(t)}$ denote the hidden state, output (action), and input (observation) vectors, respectively, at timestep $t$.}.
\begin{align*}
    h^{(t)} &= \tanh(W_hh^{(t-1)} + W_xx^{(t)} + b_h) \\
    o^{(t)} &= W_oh^{(t)} + b_o
\end{align*}

\subsection{Gated Recurrent Unit}
A GRU improves upon the vanilla RNN by learning to retain context for the next action by controlling how much of the inputs and previous hidden state to allow between timesteps \cite{gru}. GRUs have a reset gate $r$ after the previous activation to forget part of the previous state and an update gate $u$ decides how much of the next activation to use for updating$^1$. 
\begin{align*}
    u^{(t)} &= \sigma(W_{u,x}x^{(t)} + W_{u,h}h^{(t-1)} + b_u) \\
    r^{(t)} &= \sigma(W_{r,x}x^{(t)} + W_{r,h}h^{(t-1)} + b_r) \\
    h^{(t)} &= u^{(t)} \circ h^{(t-1)} + (\vec{1} - u^{(t)}) \\
    &\circ \tanh(W_{h,x}x^{(t)} + W_{h,h}h^{(t-1)} + b_h) \\
    o^{(t)} &= W_{o,h}h^{(t)} + b_o
\end{align*}

\subsection{Long Short-Term Memory}
An LSTM learns a "memory" of important locomotion context via input, forget, and output gates \cite{lstm1, lstm2}. The input gate $i$ regulates how much of the new cell state to keep, the forget gate $f$ regulates how much of the existing memory to forget, and the output gate $o$ regulates how much of the cell state should be exposed to the next layers of the network\footnote{$h^{(t)}$, $c^{(t)}$, and $x^{(t)}$ denote the hidden state, cell state, and input (observation) vectors, respectively, at timestep $t$. The output is produced with a linear mapping of $h^{(t)}$ to the output (action) vector.}.
\begin{align*}
    f^{(t)} &= \sigma(W_{f,x}x^{(t)} + W_{f,h}h^{(t-1)} + b_f) \\
    i^{(t)} &= \sigma(W_{i,x}x^{(t)} + W_{i,h}h^{(t-1)} + b_i) \\
    o^{(t)} &= \sigma(W_{o,x}x^{(t)} + W_{o,h}h^{(t-1)} + b_o) \\
    c^{(t)} &= f^{(t)} \circ c^{(t-1)} + i^{(t)} \\
    &\circ \tanh(W_{c,x}x^{(t)} + W_{c,h}h^{(t-1)} + b_c) \\
    h^{(t)} &= o^{(t)} \circ \sigma(c^{(t)})
\end{align*}

\section{Acknowledgements}
This research on modeling central pattern generators originally began as a class project for CS 229, Machine Learning. We thank our TA advisor, Mario Srouji, for giving us insight on his Structured Control Net \cite{scn}. We also thank Rachel Liu for contributing to the ideation and implementation of the RCN.

\bibliographystyle{plain}
\bibliography{main}
\end{document}